\documentclass[default,a4paper]{svproc}

\usepackage{amsmath,amssymb,amsfonts}
\usepackage{framed}
\usepackage{float}
\usepackage{algorithmic}
\usepackage{graphicx}
\usepackage{graphicx}
\usepackage{xcolor}
\usepackage[ruled,linesnumbered,vlined]{algorithm2e}
\usepackage{verbatim}
\usepackage{tikz}
\usetikzlibrary{shapes.geometric, arrows}
\usepackage{pgfplots}
\usepackage{tabto}
\usepackage{subcaption}
\usepackage{siunitx}
\usepackage{makecell}
\pgfplotsset{compat=1.16}
\SetArgSty{textnormal}

\title{
 Temporally-Sampled Efficiently Adaptive State Lattices for Autonomous Ground Robot Navigation in Partially Observed Environments
}

\author{Ashwin Satish Menon\textsuperscript{1}, Eric R. Damm\textsuperscript{1}, Eli S. Lancaster\textsuperscript{2}, Felix A. Sanchez\textsuperscript{3}, Jason M. Gregory\textsuperscript{2}, and Thomas M. Howard\textsuperscript{1,2}
\institute{University of Rochester, Rochester, NY 14627, USA, \email{amenon@ur.rochester.edu}, 
\and
    DEVCOM Army Research Lab, Adelphi, MD 20783, USA
\and
    Parsons Corporation, Chantilly, VA 20151, USA    
    }
}
\authorrunning{Menon et al.}  
\titlerunning{TSEASL for AGR Navigation in Partially Observed Environments}  

\begin{document}
\maketitle
\begin{abstract}
Due to sensor limitations, environments that off-road mobile robots operate in are often only partially observable. As the robots move throughout the environment and towards their goal, the optimal route is continuously revised as the sensors perceive new information. In traditional autonomous navigation architectures, a regional motion planner will consume the environment map and output a trajectory for the local motion planner to use as a reference. Due to the continuous revision of the regional plan guidance as a result of changing map information, the reference trajectories which are passed down to the local planner can differ significantly across sequential planning cycles. This rapidly changing guidance can result in unsafe navigation behavior, often requiring manual safety interventions during autonomous traversals in off-road environments. To remedy this problem, we propose Temporally-Sampled Efficiently Adaptive State Lattices (TSEASL), which is a regional planner arbitration architecture that considers updated and optimized versions of previously generated trajectories against the currently generated trajectory. When tested on a Clearpath Robotics Warthog Unmanned Ground Vehicle as well as real map data collected from the Warthog, results indicate that when running TSEASL, the robot did not require manual interventions in the same locations where the robot was running the baseline planner. Additionally, higher levels of planner stability were recorded with TSEASL over the baseline. The paper concludes with a discussion of further improvements to TSEASL in order to make it more generalizable to various off-road autonomy scenarios.
\end{abstract}
\section{Introduction}

%
Field robots that navigate long distances without a priori knowledge of the environment must make estimates of traversability cost through regions that are partially obscured.
These estimates are continuously revised as the environment map is populated from sensor data collected during the traversal.
A regional motion planner generates trajectories from the current state to the goal region using these partially observed environment maps.  
When traversability cost estimates are based on distance, time, and/or cost maps, two trajectories that are similar in cost can vastly differ in shape.
When regional planners are designed to output optimal trajectories, small updates to the initial state and environment map can lead to significant differences in guidance to lower-level planners.
This is problematic for mobile robot navigation as continuous changes in guidance can lead to inefficient, unsafe behavior.  

A potentially better approach allows the regional planner to consider prior trajectories, updated and potentially optimized for the current state and environment map, in addition to the optimal trajectory generated by a sampling-based or search-based motion planner.
In this paper we propose Temporally-Sampled Efficiently Adaptive State Lattices (TSEASL) for regional motion planning that arbitrates guidance to lower-level planners in partially observed environments. 

 The motivation for this contribution is based upon the notion that traditional off-road autonomy architectures are reliant on regional planners that supply optimal guidance to local planners at a fixed, regular time cycle. Local planners like Model-Predictive Path Integral control \cite{williams2016aggressive} and Receding-Horizon Model-Predictive control \cite{howard2010receding} can then be put into potentially hazardous situations because they are required to both avoid obstacles and also follow the regional plans which are subject to change regularly (and often dramatically). The possible marginal improvement in regional plan optimality can be outweighed by the worsening of overall navigation behavior.

\label{sec:introduction}

\section{Related Work}
\label{sec:related-work}


Mobile robot motion planning has undergone significant development over the past two decades. Classical planning algorithms are generally classified into two methods: probabilistic sampling \cite{lavalle1998rapidly} \cite{karaman2011sampling} \cite{kavraki_prms_1996} and deterministic graphs \cite{pivtoraiko2009differentially}.  
The stochastic nature of sampling-based approaches can lead to inefficiencies in how the state space is explored, resulting in multiple different routes to roughly the same state. 
%
%
%
Recombinant graphs make use of control sets which recombine at fixed intervals in the state space. Robot states are represented by nodes in the graph.  More recent recombinant search spaces such as the Adaptive State Lattice \cite{howard2009adaptive}, EASL \cite{hedegaard2021discrete}, and KEASL \cite{damm2023terrain}, locally optimize the location of sampled nodes in the graph using approximations of the aggregate costs of local edges to improve the relative optimality of heuristic search.  EASL and KEASL specifically discretize the state space adaptations considered, thus limiting the set of feasible motions which could arise when search is performed. By doing this, the online trajectory generation process \cite{howard2007optimal}, which can be expensive if applied extensively in large search spaces, is eliminated. 
This enables the precomputation of swaths for edge cost evaluation, resulting in paths that are of comparable cost to the ASL solutions, but are found in less time. 
Correspondingly, the EASL planner returns paths which have lower cost than a standard state lattice-based planner, but with similar computation time. 

More recently, many methods have made use of learning-based approaches to perform planning in obstacle-dense environments. Intention-Net \cite{gao2017intention} acts as a replacement for the local planning architecture, guided by a standard A* \cite{A-star} global planner and incoming perceptual information. Similarly, PRM-RL \cite{faust2018prm} used a sampling based Probabilistic Roadmap global planner \cite{kavraki_prms_1996} to inform short-range navigation tasks that are solved via reinforcement learning. Both of these approaches, along with most other existing learning methods for planning within off-road navigation, aim to replace the local planning paradigm. Machine learning-based regional and global planning methods are not well-represented in the literature. This is potentially due to the focus of the navigation problem pertaining to obstacle avoidance and velocity planning -- subproblems which are commonly handled by local planners such as MPPI \cite{williams2016aggressive}. While MPPI generates thousands of rollouts and compares their costs in order to weight an optimal solution, the methodology is different than the one in this paper since we consider the costs of previous trajectories to select the one that prioritizes the greatest amount of plan stability. Put simply, MPPI combines thousands of trajectories with varying weights to yield a single one, while our approach selects an existing trajectory from a list of previously generated ones, subject to the constraints of the updated map. 
The aforementioned EASL approach was further extended into HAEASL \cite{menon2024homotopy}, which was formulated to return multiple solutions, each within a distinct homotopy class. This reformulation enabled the robot to consider trajectories which were not necessarily globally optimal, but did not oscillate around obstacles due to homotopy class switching on new planning cycles. The notion that suboptimal motion plans can be used to provide guidance that enables optimal navigation was explored with HAEASL by sampling motions in the spatial dimension. The work in this paper aims to solve the same problem but by sampling motions in the temporal dimension. As such, the contributions of this paper are as follows:
\begin{itemize}
	\item Temporally Sampled EASL, the first motion planner which uses an arbitration architecture that provides local planner guidance based on a comparison of relative plan costs to improve overall navigation performance using the EASL Search Space
	\item Evaluation of TSEASL supplying regional plan guidance for mobile robot navigation using data collected from a physical robot in an off-road environment subject to resource, temporal, and kinodynamic constraints.
	\item Demonstration of TSEASL running live on a Clearpath Robotics Warthog Unmanned Ground Vehicle (UGV).
\end{itemize}

\section{Technical Approach}
\label{sec:technical-approach}

\begin{figure*}[htbp!]
	\begin{center}
	\includegraphics[width=0.9\textwidth]{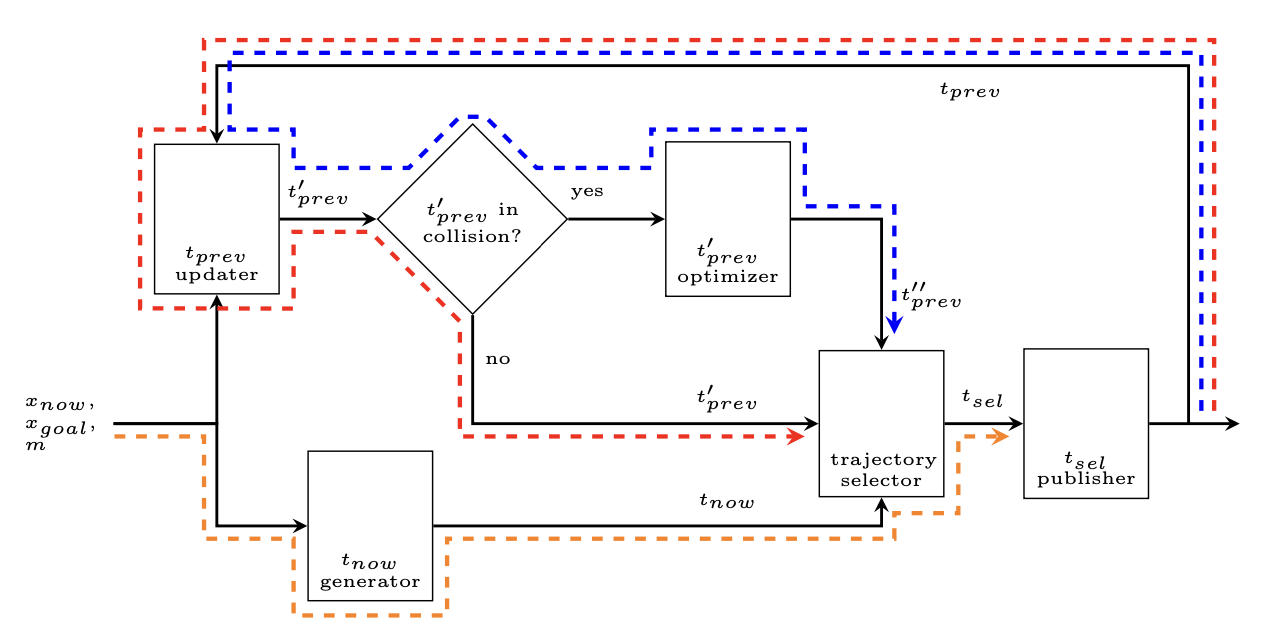}
	\end{center}
	\caption{The proposed TSEASL architecture for regional planning.}
	\label{fig:tseasl_flowchart}
\end{figure*}

The proposed architecture for TSEASL is illustrated in Figure \ref{fig:tseasl_flowchart}.  
At the center of this architecture is a block titled \textit{trajectory} \textit{selector}.
This block decides which of several possible trajectories to pass onto the local planner.
The inputs to this architecture are the current state $x_{now}$, the goal region $x_{goal}$, and the environment map $m$.  
On the first cycle of this architecture, there is no previous trajectory $t_{prev}$ so the inputs are processed by the state lattice-based motion planner ``$t_{now}$ generator''.
The trajectory selector has only this trajectory to choose from, so it is passed onto the local planner.
This route of information is illustrated as an orange dashed line.
The published trajectory now becomes $t_{prev}$ as it is the trajectory generated from the previous planning cycle.
Trajectory $t_{prev}$ is trimmed to its nearest state to $x_{now}$ and updated based on the environment map $m$ to become $t_{prev}'$.
That information now passes one of two ways.
The first route is illustrated as a red dashed line.
In this route, the updated trajectory $t_{prev}'$ is determined not to collide with any obstacles. 
The trajectory selector now compares $t_{prev}'$ to $t_{now}$ and chooses one of the two to pass onto the local planner.
A weight $\alpha$ is used to instill a bias for previously published trajectories where $\alpha < 1$.
If the cost of trajectory $t_{now} < \alpha t_{prev}'$, trajectory $t_{now}$ is published to the local planner and becomes the new $t_{prev}$.
Otherwise $t_{prev}'$, which is the same shape but with updated velocities and duration based on the updated state $x_{now}$ and environment map $m$, is published to the local planner.
The second possible route is illustrated as a blue dashed line.
In this route, the updated trajectory $t_{prev}'$ is determined to be in collision with an obstacle in $m$.  

\begin{figure*}[htbp!]
\begin{center}
\includegraphics[width=0.8\textwidth]{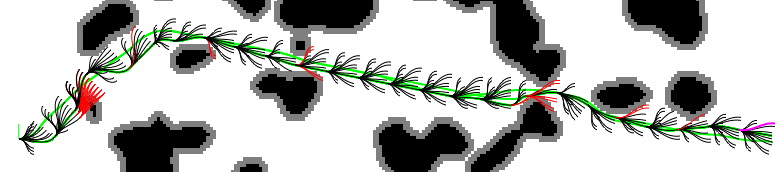}
\end{center}
\caption{An illustration of the search space explored by ARA* in a trajectory-aligned state lattice used in the $t_{prev}'$ optimizer step of the architecture proposed in Figure \ref{fig:tseasl_flowchart}. The map was synthetically generated using Perlin noise \cite{perlin2002improving}.}
\label{fig:node_optimizer}
\end{figure*}

Since it may be desirable to make small adjustments to the prior trajectory rather than just take the newly generated trajectory $t_{now}$, we pass this trajectory into a planner ``$t_{prev}'$ optimizer'' that samples nodes laterally around the updated previous trajectory $t_{prev}'$.
The rest of $t_{prev}'$ may not need to be regenerated if only one part of it is in collision. By only repairing parts of the trajectory that are absolutely needed, the node optimizer often takes less time to compute than finding a new trajectory in the updated map. The dense lateral samples are used in order to find nodes that are not in collision with the updated map's obstacles. Figure \ref{fig:node_optimizer} displays this process occurring over a full trajectory in a synthetically generated map using Perlin noise \cite{perlin2002improving}. The red edges are lateral samples which are in collision.

Here, an initial trajectory with a duration of 35.69 seconds is generated in 0.46 seconds using Kinodynamic EASL \cite{damm2023terrain} with Anytime Repairing A* (ARA*) \cite{ARA}.  ARA* is then used with a state lattice aligned with the original trajectory with states sampled laterally from the original trajectory at a regular interval.  
The optimized trajectory in this scenario yielded a motion with a duration of 29.05 seconds in 0.12 seconds, a reduction of approximately 19\%. It is also important to note that the node optimized trajectory was found in approximately 26\% of the time it took to compute the original trajectory, further motivating the utility of local node repairs instead of replanning an entirely new trajectory.
The optimized trajectory $t_{prev}''$ is now compared against the newly generated trajectory $t_{now}$ using the relation $t_{now} < \alpha t_{prev}''$.
If the cost of trajectory $t_{now} < \alpha t_{prev}''$, trajectory $t_{now}$ is published to the local planner and becomes the new $t_{prev}$.
Otherwise $t_{prev}''$, which is approximately but not exactly the same shape as $t_{prev}'$, is published to the local planner.




\section{Experiments and Results}
\label{sec:experimental-design}

\begin{figure}[ht!]
	\centering
	\includegraphics[width=.64\linewidth]{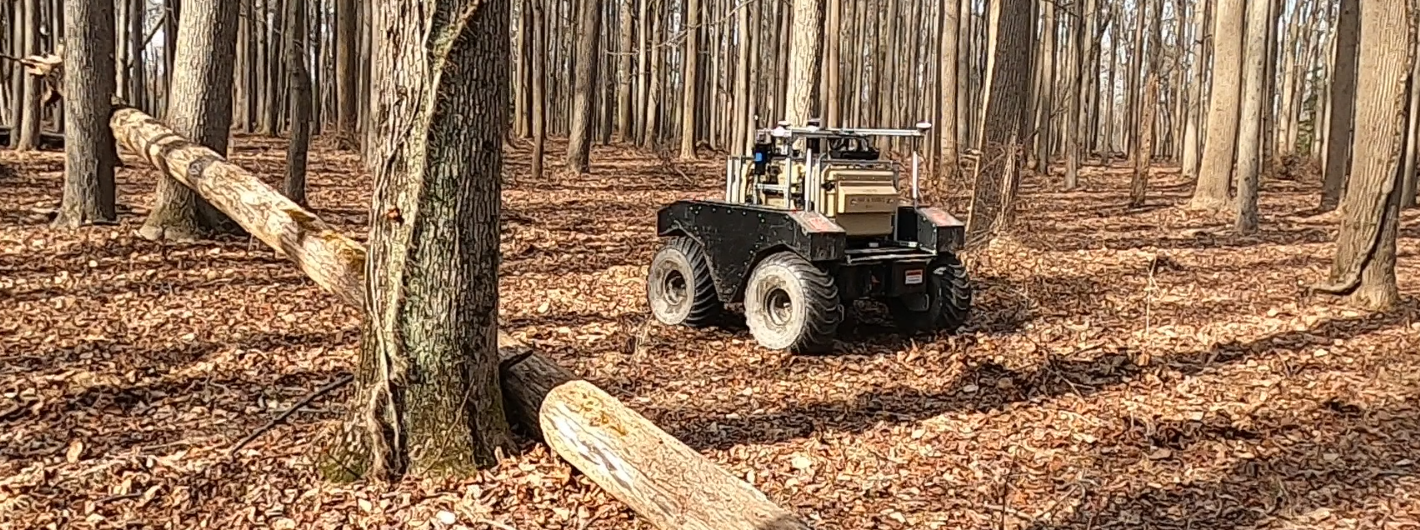}
	\caption{The forested environment TSEASL was tested in to compare its performance against the baseline.}
	\label{fig:test_scenarios}
\end{figure} 

\begin{figure*}[htbp]
	\centering
	\textcolor{orange}{\frame{\includegraphics[height=0.17\textwidth]{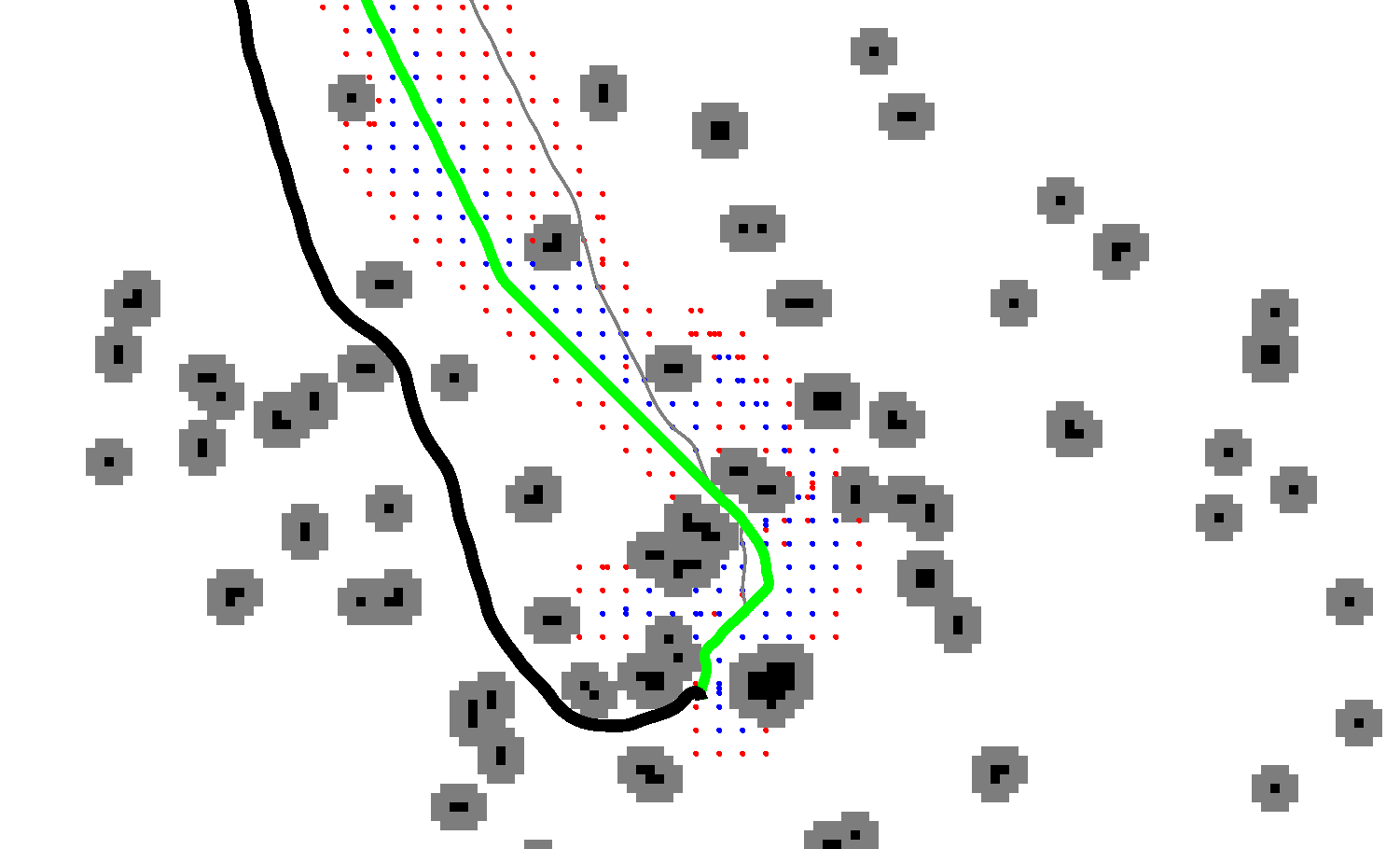}}}
	\textcolor{orange}{\frame{\includegraphics[height=0.17\textwidth]{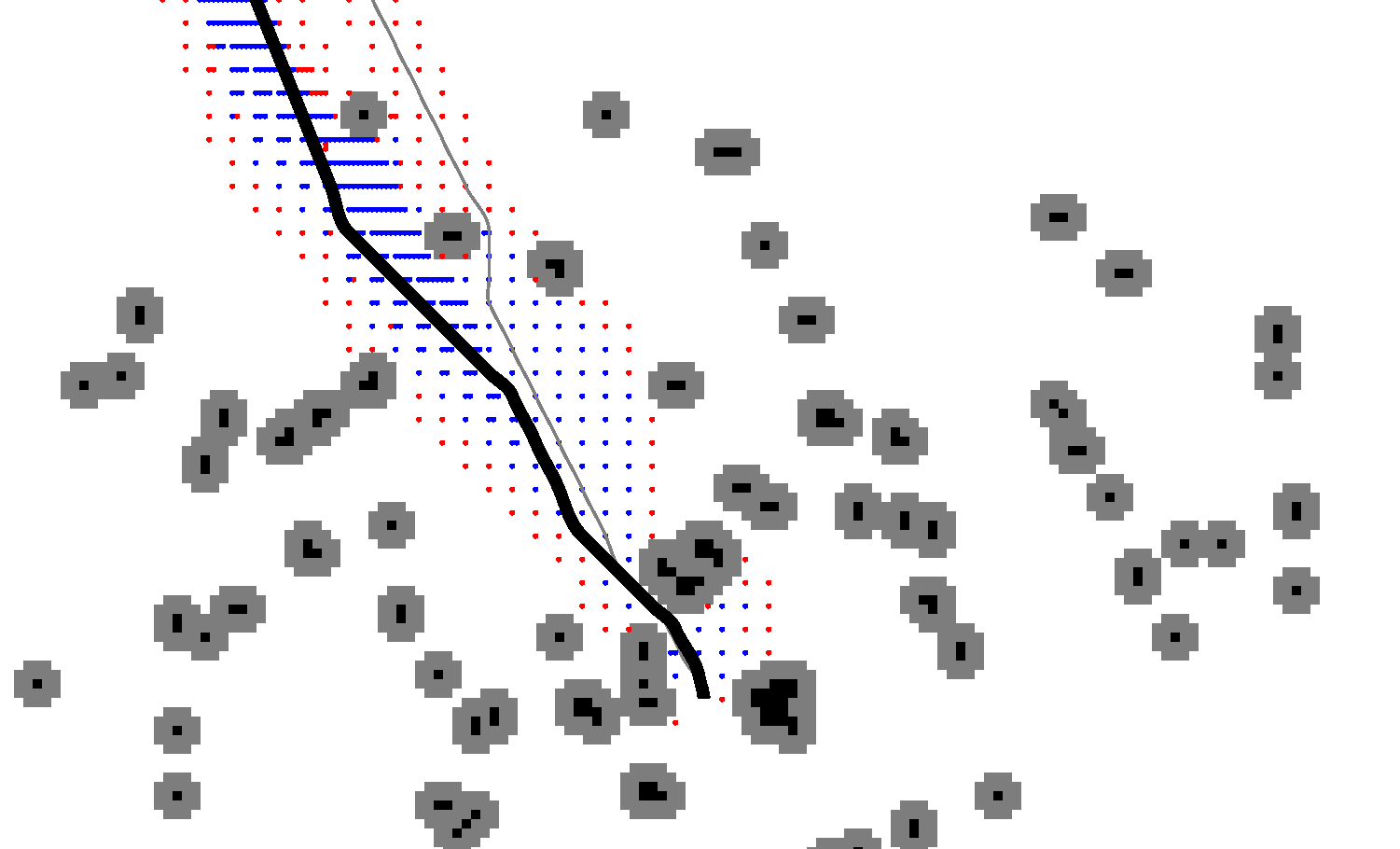}}}
	\caption{Two consecutive planning cycles illustrating when $t_{now}$ has a sufficiently cheaper cost than $t_{prev}'$. The cost of $t_{prev}'$ was 56.83 seconds, while the cost of $t_{now}$ in the right figure was 52.87 seconds. Since this cost difference is $>$ 5\% (approximately 6.98\%), TSEASL decides that it is worth it to switch its $t_{sel}$ to pass down to the local planner. The orange borders indicate that this example took the orange decision line in Figure \ref{fig:tseasl_flowchart}.}
	\label{fig:tnow}
\end{figure*}

\begin{figure*}[htbp]
	\centering
	\textcolor{red}{\frame{\includegraphics[height=0.17\textwidth]{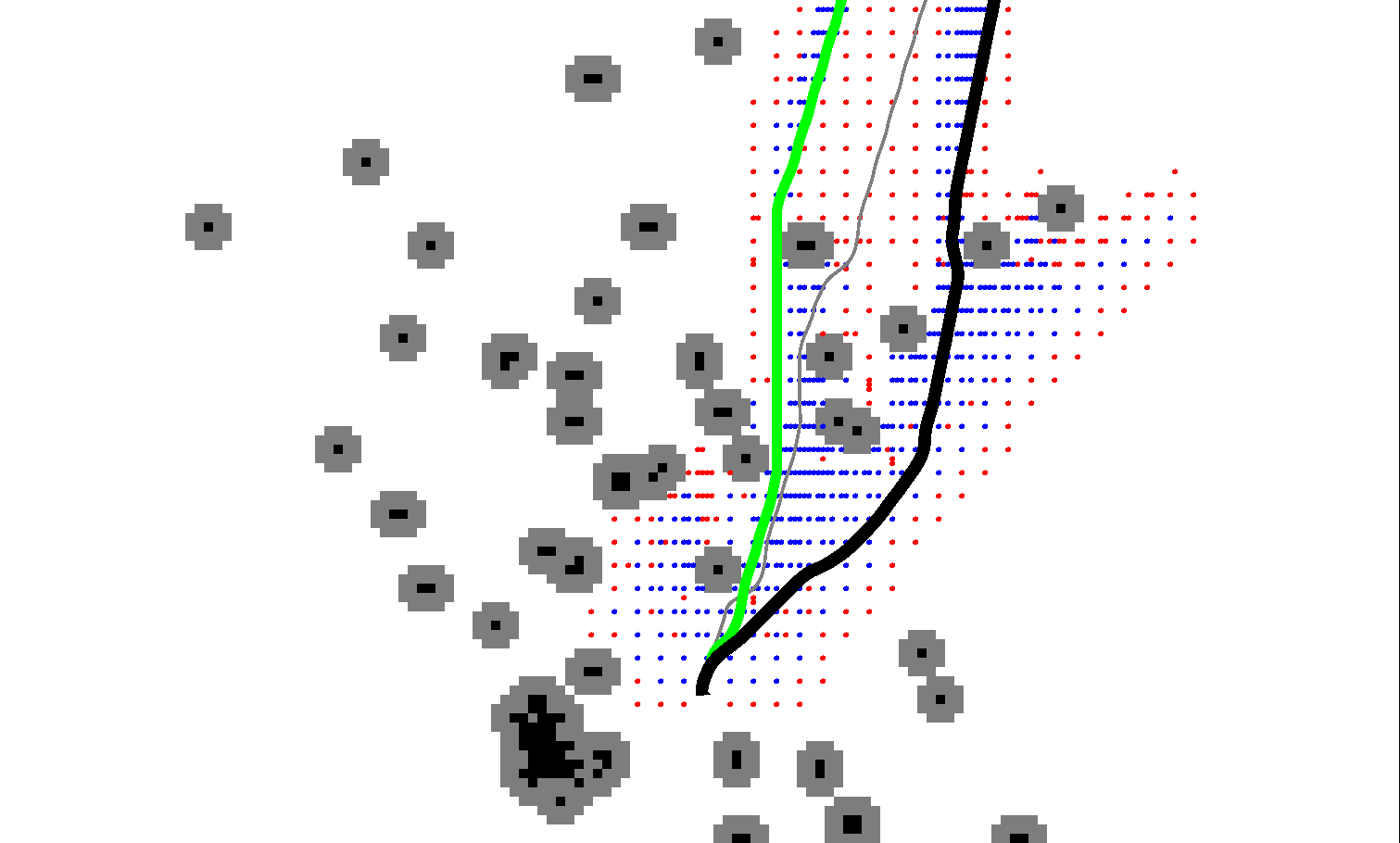}}}
	\textcolor{red}{\frame{\includegraphics[height=0.17\textwidth]{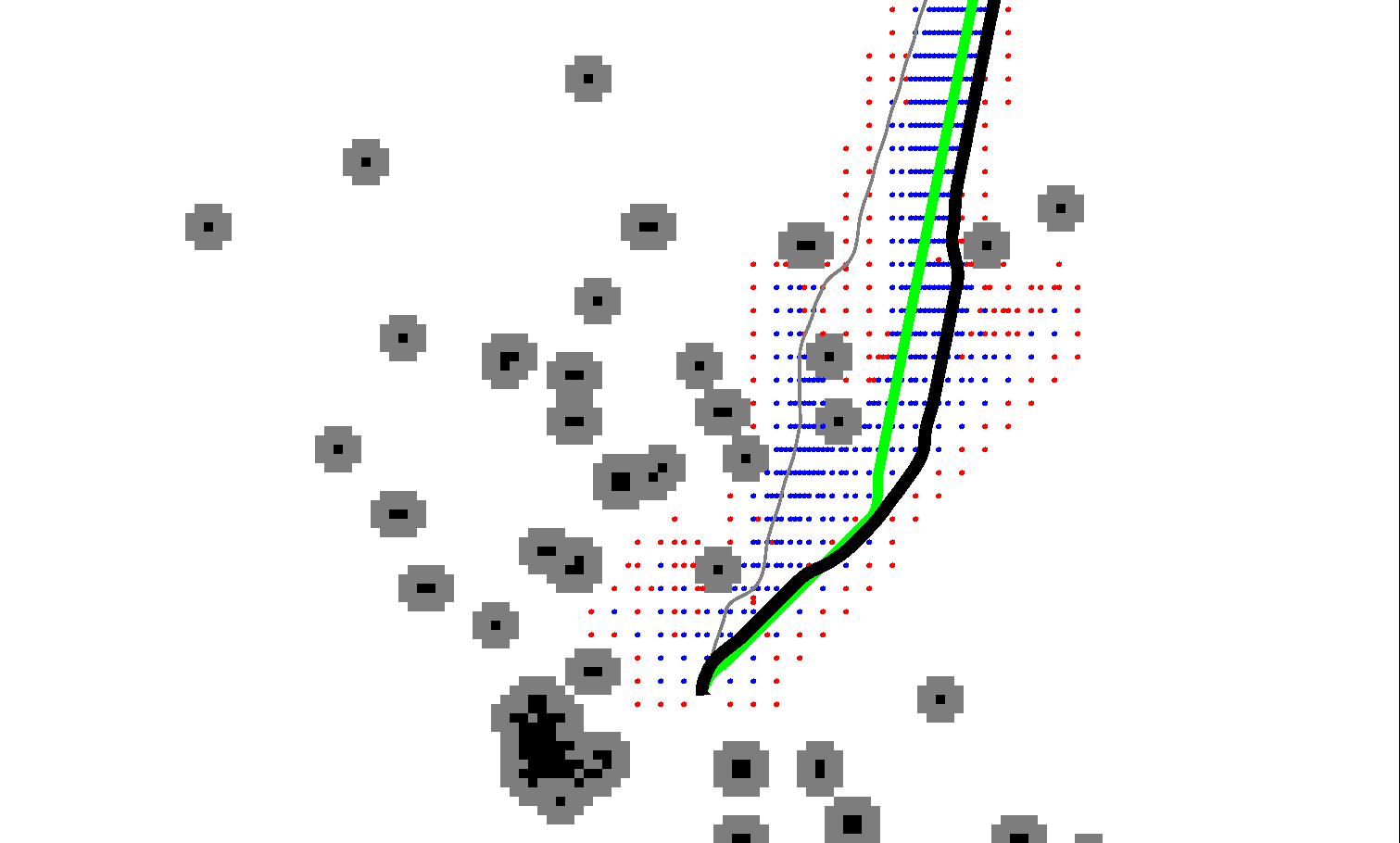}}}
	\textcolor{red}{\frame{\includegraphics[height=0.17\textwidth]{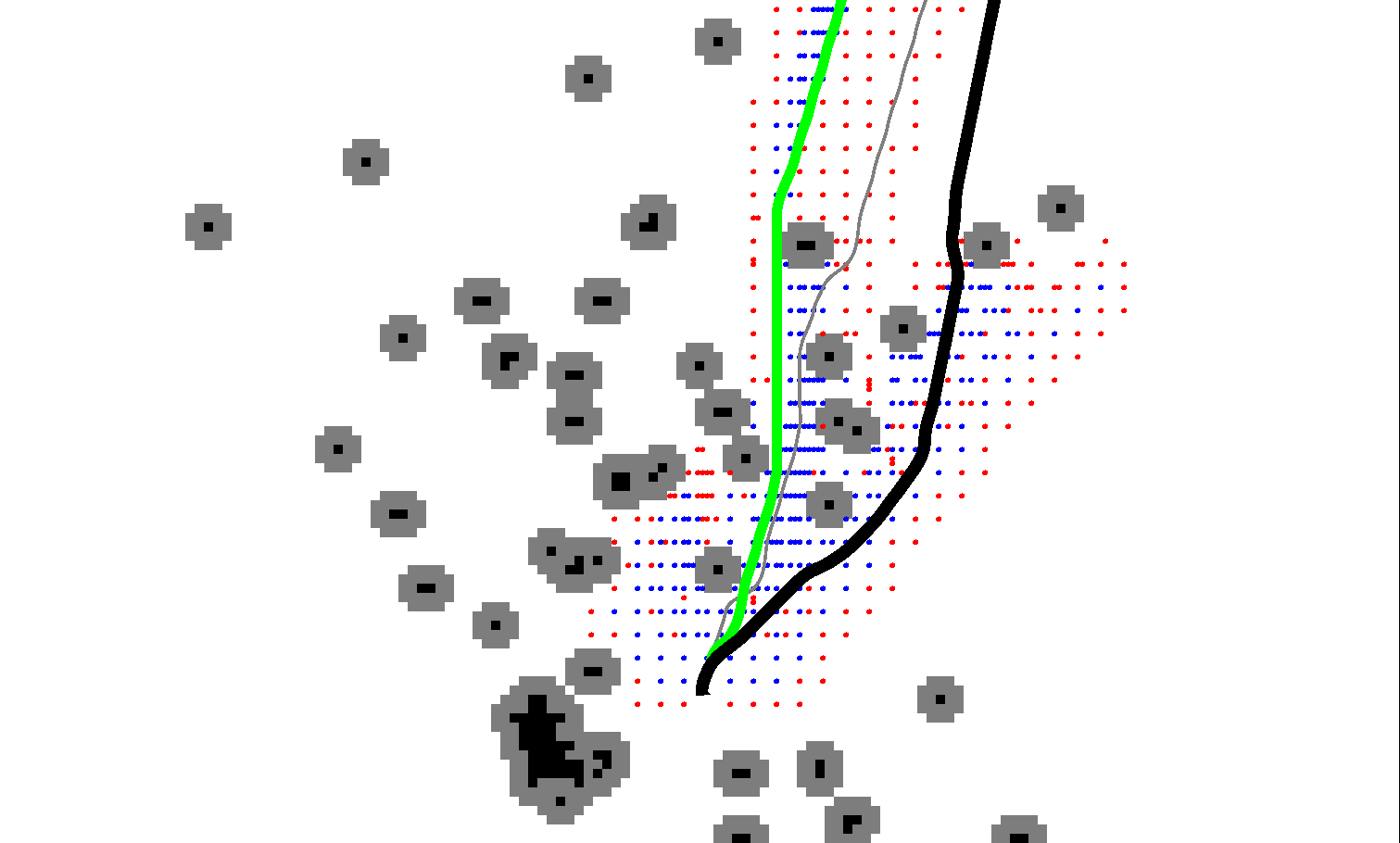}}}
	\caption{Three consecutive planning cycles when $t_{now}$ does not have a sufficiently cheaper cost than $t_{prev}'$. The cost of $t_{now}$ changed from 48.24 seconds, to 48.28 seconds, back to 48.24 seconds, while the cost of $t_{prev}'$ in all three figures stayed at 48.70 seconds. The cost difference between $t_{now}$ and $t_{prev}'$ was never $>$ 5\%, so $t_{sel}$ never changes. The red borders mean this example took the red decision line in Figure \ref{fig:tseasl_flowchart}.}
	\label{fig:tprev_prime}
\end{figure*}

\begin{figure*}[htbp]
	\centering
	\textcolor{blue}{\frame{\includegraphics[height=0.17\textwidth]{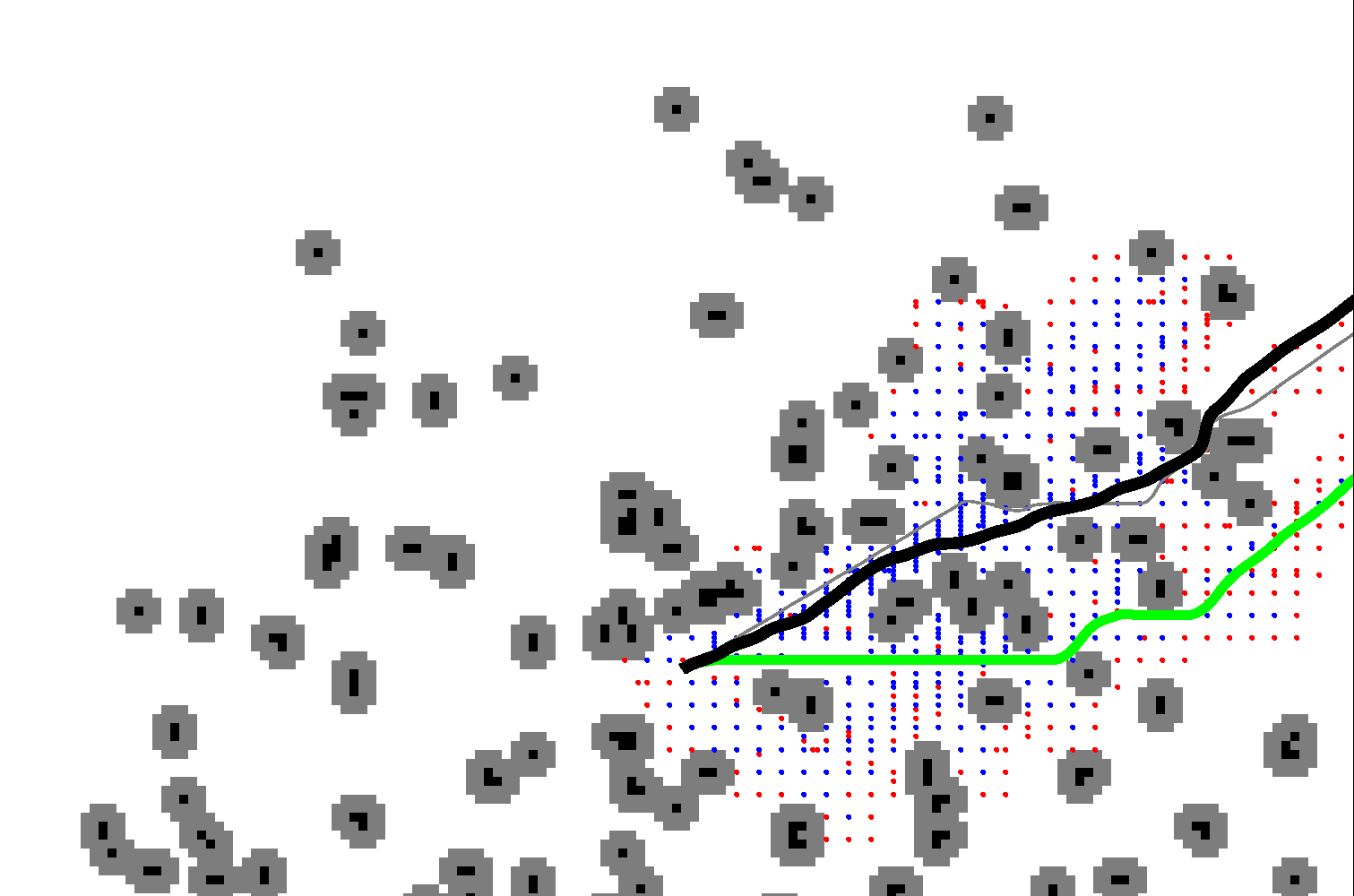}}}
	\textcolor{blue}{\frame{\includegraphics[height=0.17\textwidth]{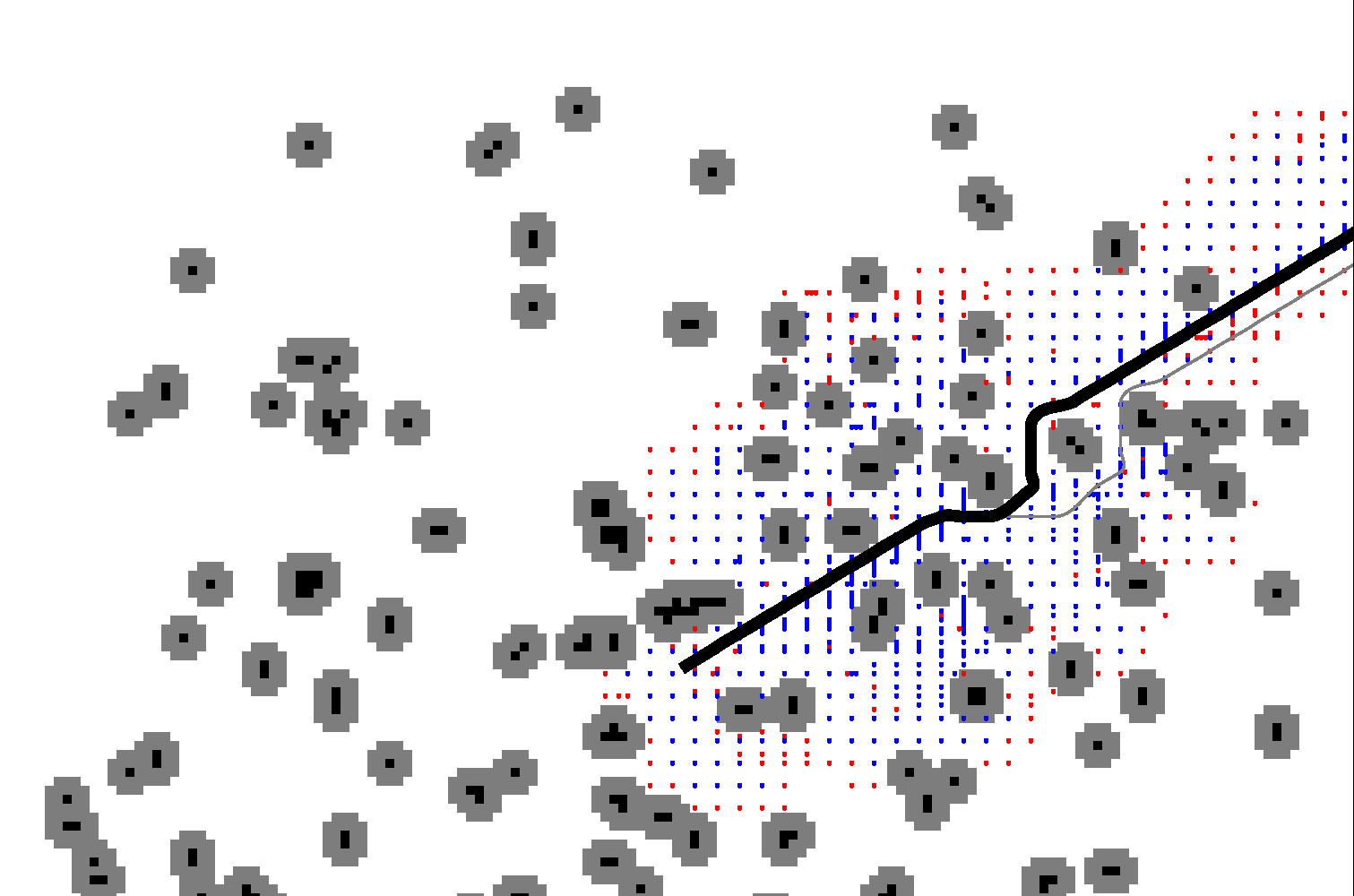}}}
	\textcolor{blue}{\frame{\includegraphics[height=0.17\textwidth]{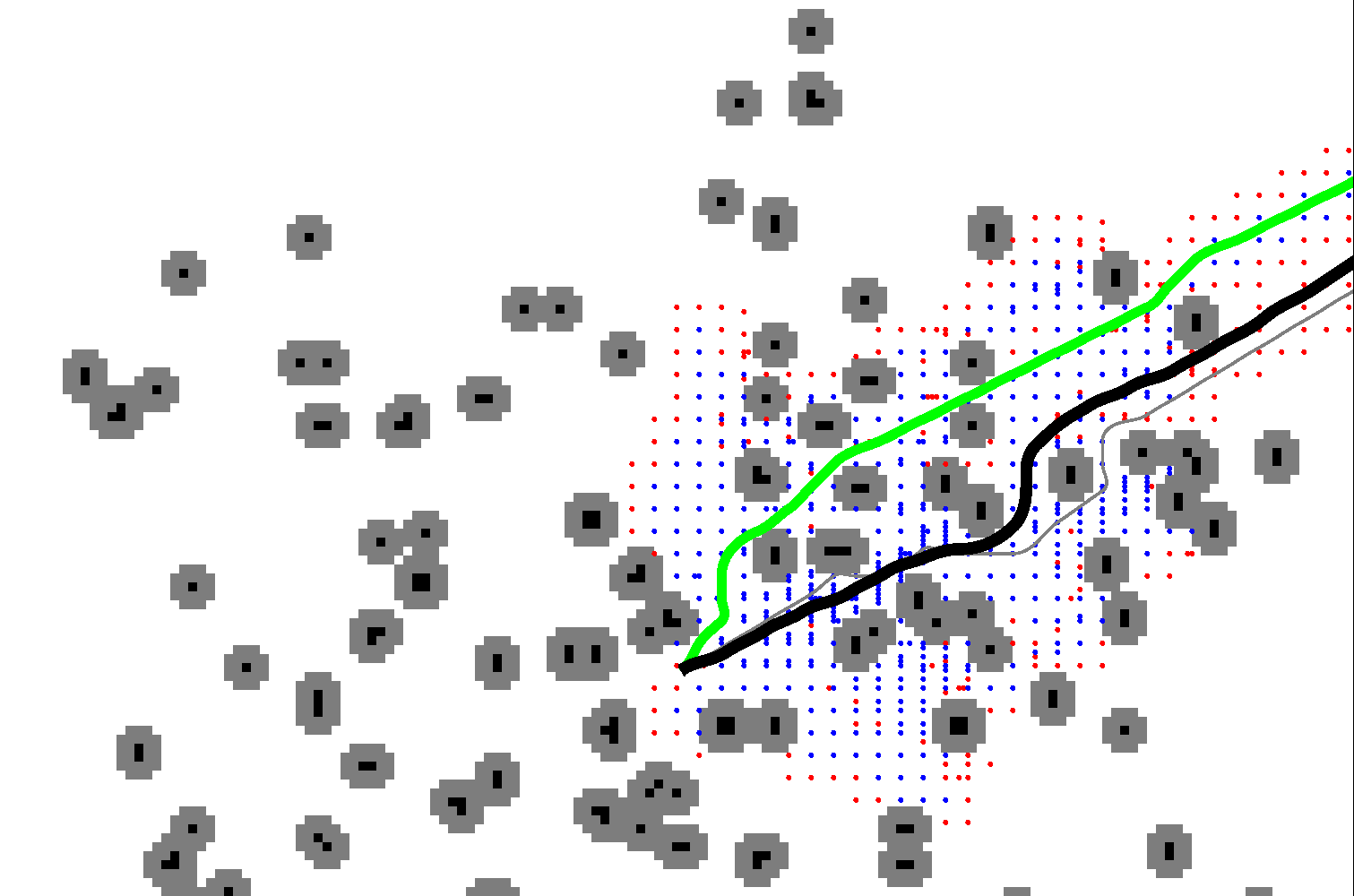}}}
	\caption{Left: TSEASL performs a node optimization on $t_{prev}'$ to turn it into $t_{prev}''$  so it can pass through the top right of the map. Middle: TSEASL attempts to node optimize further, but the top right of the map is no longer traversable due to new map information. TSEASL falls back on $t_{now}$ as the trajectory it supplies as $t_{sel}$. Right: the green trajectory, $t_{now}$, is KEASL's solution with a cost of 40.8236 seconds. $t_{prev}'$ has a cost of 42.01 seconds, but it is in collision with obstacles in the updated map. 
	The blue borders mean this example took the blue decision line in Figure \ref{fig:tseasl_flowchart}.}
	\label{fig:tprev_doubleprime}
\end{figure*}

\noindent \textbf{Case Study Experiments:} TSEASL ran on a Clearpath Robotics Warthog Unmanned Ground Vehicle (UGV) in an off-road forested environment, as shown in Figure \ref{fig:test_scenarios}. The environment is laden with dense crops of trees and other vegetation serving as obstacles.%
During the forest autonomy runs, $\alpha$ was set to $0.95$. When compared to baseline KEASL, but with all other elements of the stack maintained, TSEASL exhibited fewer interventions due to navigation performance over an approximately 700 meter forest course. This was due in large part to an increase in plan stability observed during the preliminary experiments, as shown in Figure \ref{fig:tnow} through Figure \ref{fig:tprev_doubleprime}. Each figure demonstrates a different output for the trajectory selector shown in Figure \ref{fig:tseasl_flowchart}. For all three of these figures, the black trajectory represents the output of TSEASL while the green trajectory represents the output of KEASL. In the images where no green trajectory is visible, TSEASL and KEASL generated the same output. The black map cells are obstacles, while the gray map cells are the expanded obstacles to respect the footprint of the robot. White cells are free space. Finally, the blue and red dots represent closed and open lists, respectively, to indicate which nodes in the graph have been explored and which nodes are yet to be explored.

Figure \ref{fig:tnow} illustrates an example of TSEASL deciding to use the newest generated trajectory, termed $t_{now}$. Since the cost of $t_{now}$ is more than 5\% lower than the cost of $t_{prev}'$, the selected trajectory, $t_{sel}$, is chosen as $t_{now}$. 
Figure \ref{fig:tprev_prime} shows a sequence when KEASL's trajectories are changing to reflect the updating map information, but TSEASL's trajectory does not change because the newest generated trajectory is not sufficiently better ($>$5\%) than TSEASL's $t_{prev}'$. As a result, TSEASL's $t_{sel}$ stays consistent. 
Figure \ref{fig:tprev_doubleprime} demonstrates when $t_{prev}'$ turns into $t_{prev}''$ due to the unoptimized trajectory being in collision with the updated map. Thus, in the leftmost and rightmost images of this figure, TSEASL's $t_{sel}$ are published from $t_{prev}''$ because they both undergo node optimization. The middle image in this figure occurs because TSEASL fails to node optimize due to the area near the top right of the map becoming untraversable. In the right image, $t_{prev}'$ undergoes node optimization and becomes $t_{prev}''$. Since $t_{now}$ is not sufficiently cheaper than $t_{prev}''$, $t_{prev}''$ gets published as $t_{sel}$.

\noindent \textbf{Full Scale Experiments:} 

The above experiments exhibit certain advantages that TSEASL offers over the baseline, single planner approach. However, these advantages are only demonstrated in an isolated manner. To more thoroughly assess the benefits of TSEASL over the baseline, quantitative experiments are performed across 1,747 examples of planning problems taken from the off-road environment which our approach was tested in.

This experiment was designed to measure the impact of the aforementioned $\alpha$ value on the amount of plan stability that TSEASL supplied to the navigation architecture. Plan stability was measured as the Modified Hausdorff Distance (MHD) \cite{Dubuisson1994AMH}:

\begin{equation} \label{eq:1}
h(A,B)= \frac{1}{|A|}\sum_{a_i \in A}^{}(\min_{b_j\in B} d(a_i,b_j)),
\end{equation}

Equation \ref{eq:1} was used as a numerical way to calculate the path deviation between successive trajectories generated by the regional planner. A MHD of zero indicates that the two trajectories did not deviate from each other. The distribution of MHDs were surveyed and compared across the baseline KEASL planner and six versions of TSEASL, each with an $\alpha$ value ranging from 0.95 to 0.999. Recall that TSEASL with an $\alpha$ value of 1.00 is functionally the same algorithm as the baseline planner, just with the addition of the node optimizer.

\begin{figure*}[htbp]
	\centering
	\includegraphics[height=0.45\textwidth]{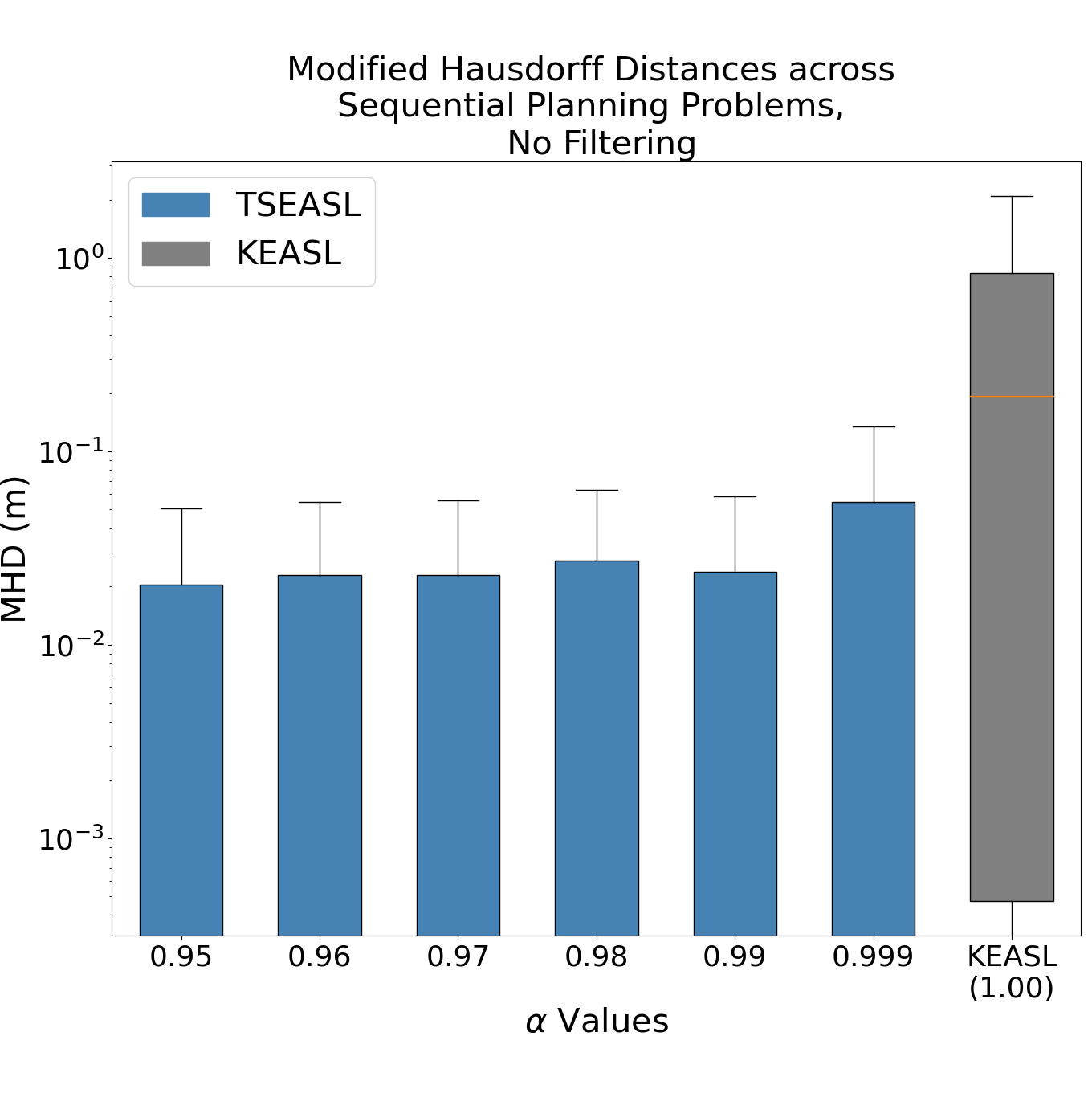}
	\includegraphics[height=0.45\textwidth]{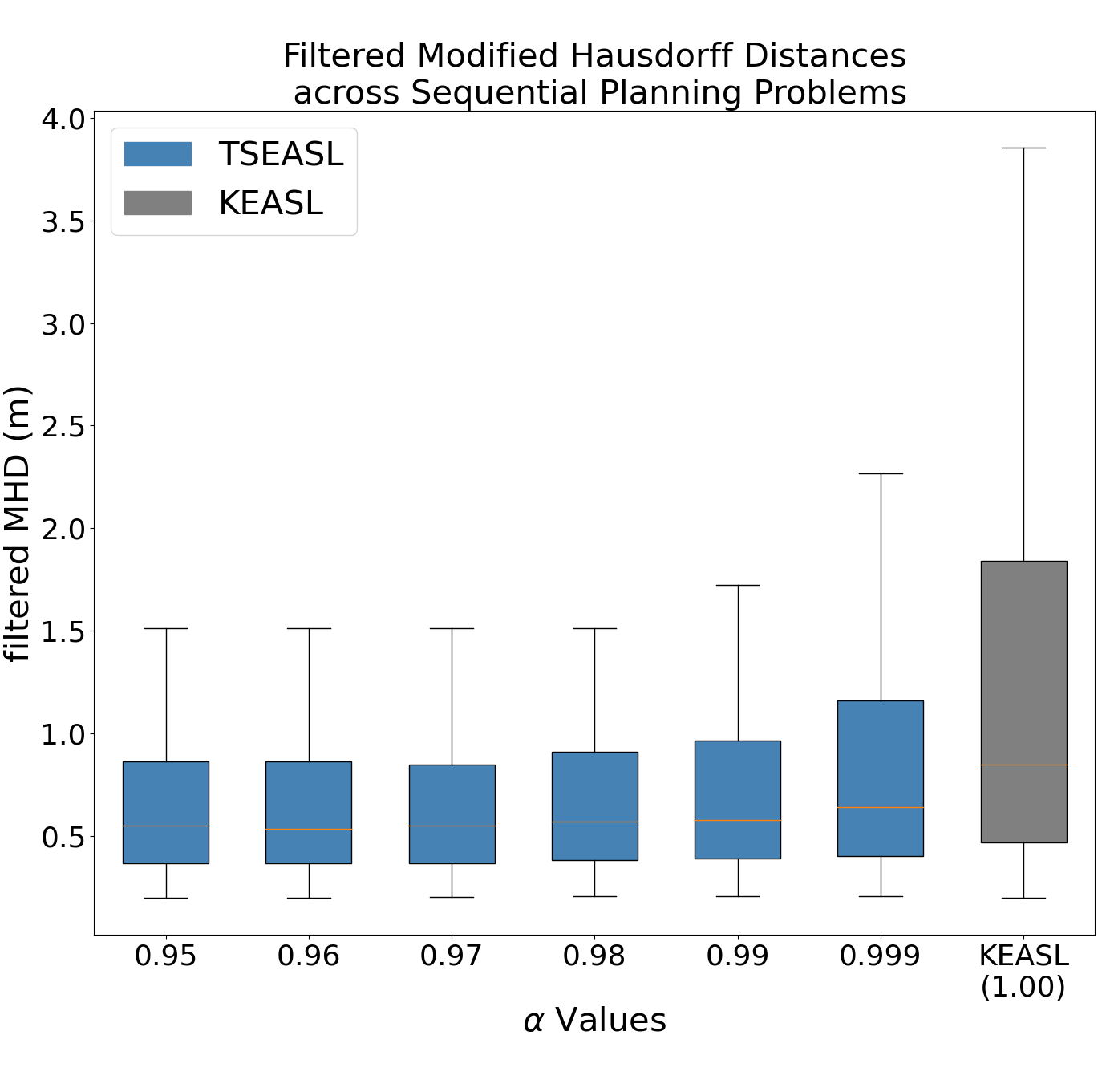}
	\caption{A comparison of Modified Hausdorff Distances across seven different planner setups: KEASL and six different versions of TSEASL, where each one uses a different $\alpha$ value.}
	\label{fig:tseasl_mhd}
\end{figure*}

Figure \ref{fig:tseasl_mhd} shows two different interpretations of the same results. The left box plot illustrates the distribution of path deviations across six iterations of TSEASL as well as the baseline KEASL planner. Note that the axis in this plot is logarithmic, since the majority of the TSEASL MHDs tended to be zero. The right box plot is the same data, but filtered to exclude MHDs which were below 0.2m. This filter threshold was chosen because it was the resolution of the cost maps that all planners performed search through. As expected, the interquartile range of KEASL's MHDs spanned larger values, indicating higher amounts of path deviation throughout all of the planning problems. Irrespective of if the filter was used or not, TSEASL's path deviations were less than KEASL's path deviations on average.

\begin{table}
\centering
\caption{\centering{Filtered Modified Hausdorff Distance by Planner Configuration}}
\resizebox{\textwidth}{!}{%
\begin{tabular}{| >{\centering\arraybackslash}p{0.3\textwidth} | >{\centering\arraybackslash}p{0.3\textwidth} | >{\centering\arraybackslash}p{0.3\textwidth} | >{\centering\arraybackslash}p{0.3\textwidth} |} \hline
Planner Configuration  & Mean ($\mu$) & \# filtered values & \# MHD zeros  \\ \hline
\begin{tabular}[c]{@{}l@{}}TSEASL, $\alpha$ = 0.95 \\\end{tabular} & 1.001   & 332           & 997                   \\ \hline
TSEASL, $\alpha$ = 0.96                & 1.020   & 335           & 988                   \\ \hline
TSEASL, $\alpha$ = 0.97                & 1.033   & 340           & 985                   \\ \hline
TSEASL, $\alpha$ = 0.98                     & 1.079   & 350           & 983                   \\ \hline
TSEASL, $\alpha$ = 0.99     & 1.058   & 336           & 976                   \\ \hline
TSEASL, $\alpha$ = 0.999         & 1.182   & 378           & 953                   \\ \hline
KEASL                            & 1.365   & 868           & 60               \\    \hline
\end{tabular}}
\label{tab:mhds}
\end{table}

This data is further expressed in Table \ref{tab:mhds}, where the mean of each planner configuration is shown when excluding MHD values below 0.2m. The third column shows the number of planning problems out of 1,747 which met the filtering criteria, while the fourth column shows the number of sequentially generated trajectories which yielded MHDs of zero. The dramatic drop from the TSEASL values to the 60 for KEASL is because KEASL returned MHDs which were often inconsequentially small due to incremental changes in planning time cost. TSEASL, however, did not exhibit this behavior, but rather reused a previous trajectory when it was feasible. Even with the small MHD values filtered out, the mean values for KEASL were still higher than all of the TSEASL variations' mean values, indicating that TSEASL demonstrated an improvement in sequential plan stability.

\begin{figure*}[htbp]
	\centering
	\includegraphics[height=0.21\textwidth]{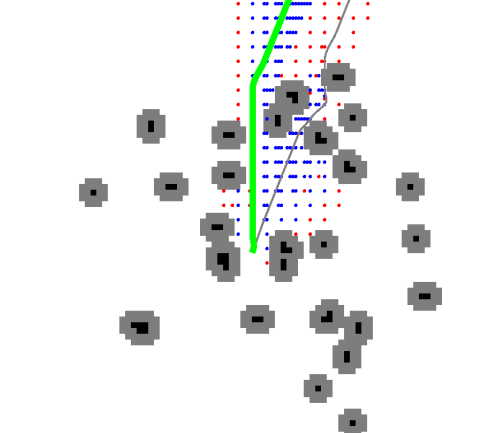}
	\includegraphics[height=0.21\textwidth]{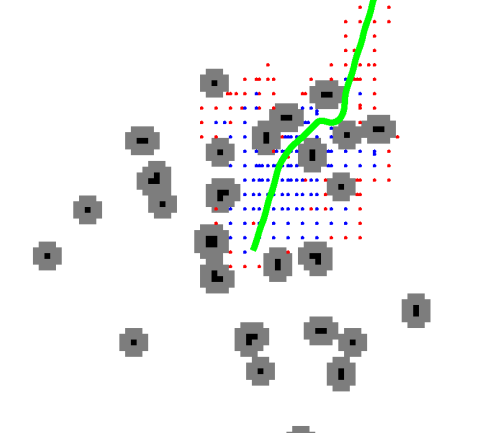}
	\includegraphics[height=0.21\textwidth]{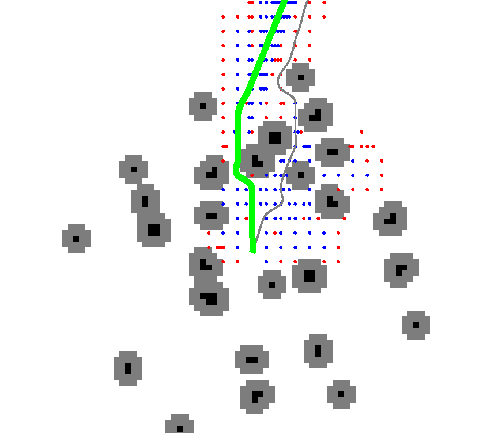}
	\includegraphics[height=0.21\textwidth]{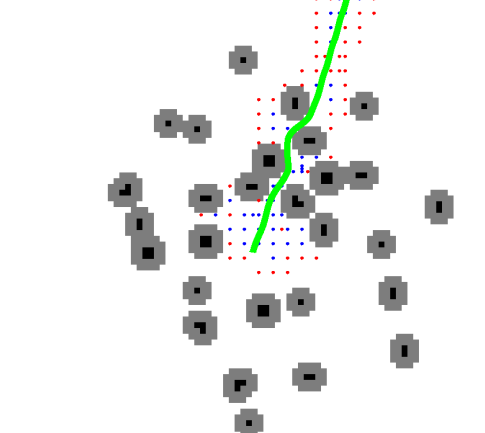}
	
	\vspace{0.5cm}
	
	\includegraphics[height=0.21\textwidth]{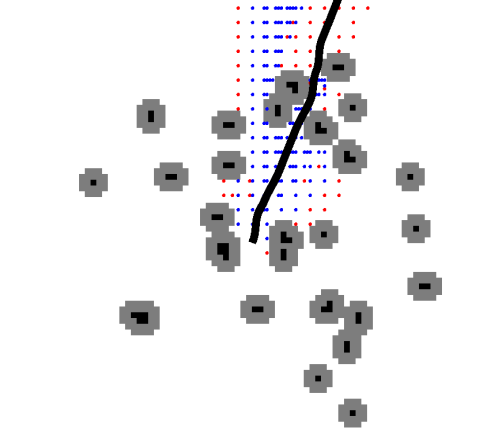}
	\includegraphics[height=0.21\textwidth]{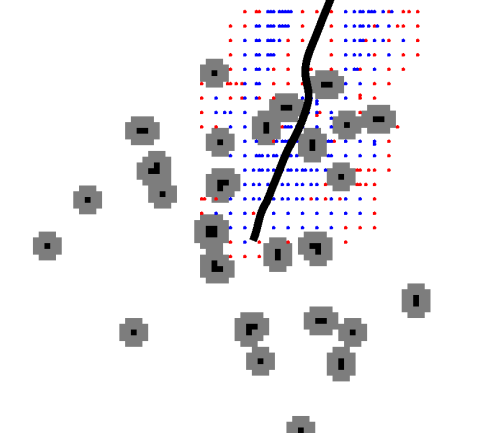}
	\includegraphics[height=0.21\textwidth]{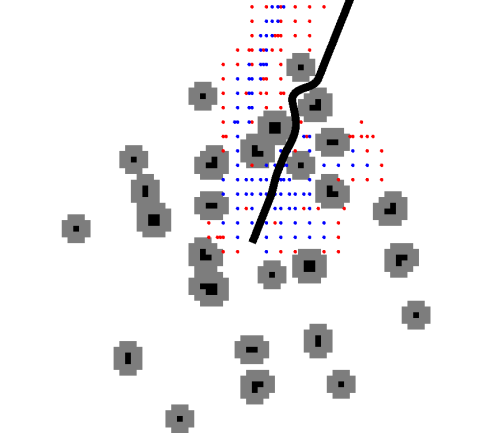}
	\includegraphics[height=0.21\textwidth]{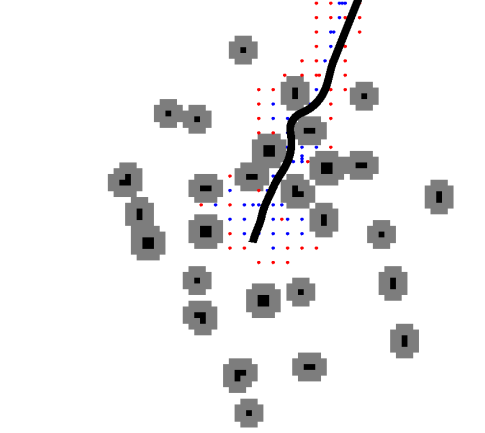}
	\caption{Four sequential planning cycles with the baseline KEASL planner (top) and the TSEASL  planner with an $\alpha$ value of 0.95 (bottom). Due to sensing limitations, KEASL frequently updates its regional plans (green) in a way which oscillates back and forth across obstacles, resulting in a safety intervention in the top rightmost image. TSEASL's regional plans (black) induce greater plan stability and bypasses the need for intervention. }
	\label{fig:planners_on_robot}
\end{figure*}

Figure \ref{fig:planners_on_robot} illustrates two live robot runs in the same dense forest on the Clearpath Warthog UGV with two distinct regional planner setups.The top row shows four sequential planning cycles when running the baseline Kinodynamic EASL (KEASL) motion planner. The regional plan continuously switches between either side of a crop of trees because it is finding slightly lower cost plans, causing the robot to drive progressively closer to the obstacles it was trying to avoid in the first place. In the second image of the top row, KEASL's solution has a duration cost of 45.254 seconds. In the third image of the top row, KEASL's solution has a duration cost of 44.491 seconds. Note that this is a 1.7\% reduction in trajectory cost, which is less than the 5\% threshold that TSEASL used for this experiment. As a result, at the same point in the environment, TSEASL did not oscillate between planning cycles. Instead, it performed a node optimization and was able to generate a solution in the same homotopy class. For KEASL, ultimately a manual safety intervention was necessitated to avoid a collision with the trees. This is contrasted with TSEASL in the bottom row which did not require a safety intervention. The regional plan that is passed down to the local planner for path following is shown to be more stable, and as a result, offers improved overall navigation. The Warthog is able to autonomously traverse this area without the need for a manual safety intervention, and it continues the mission. It is also worth noting that the Modified Hausdorff Distance for these sequential planning cycles with KEASL was 3.64m, while TSEASL's at the same point was 0.81m. This supports the notion that lower MHD values, which induce greater planner stability, can improve navigation performance.

\section{Discussion}
While the approach presented here has shown initial success with increasing plan stability to result in more robust mobile robot navigation, there are multiple directions of future work that we would like to explore. In Section \ref{sec:experimental-design}, we evaluate the effect of varying the cost threshold value at which TSEASL decides a newly generated trajectory is sufficiently better than any previous candidate trajectories. The purpose of changing this value is to study its impact on plan stability, and consequently, navigation performance.
While this is a good first step, it does not account for rapidly changing map context which the robot may encounter in many off-road environments. As such, we are interested in developing a version of TSEASL which dynamically selects the optimal cost threshold value, subject to the robot's surroundings. There may be situations, particularly in sparse, open fields, where not having a cost threshold value at all is preferable because the risk of obstacle collision is low. In these instances, the motivating problem of oscillatory regional planner guidance would rarely present itself, and the TSEASL arbitration architecture would not be necessary. Instead, running the baseline KEASL planner in the overall autonomy stack would yield equivalent navigation performance without the computational overhead of having to consider previous trajectories. 

Another area for further development was highlighted when running TSEASL on the Warthog in the forest. Due to implicit path following error from the local planner, there were occasions when the robot veered multiple meters away from the reference trajectory which TSEASL passed down. In some of these instances, due to curvature constraints in the robot's motion model, the robot was too far from the reference trajectory to continue following it. Since the cost of the selected trajectory was significantly lower than the replanned candidate trajectory when the robot was close to the mission goal, TSEASL would never supply the local planner with a reference trajectory it was capable of tracking. Instead, it would continue to supply the lower-cost trajectory that the robot could not recover back to. A solution that we aim to implement is a distance threshold value. If the robot is far enough away from the selected trajectory, the TSEASL arbiter will select $t_{now}$ instead of passing down $t_{prev}$ as it normally would, even if $t_{now}$ is not sufficiently better than $t_{prev}$ from a path cost perspective.
It is also worth noting that while TSEASL is built upon the KEASL search space, it is not restricted to using KEASL as a planner because at its core, TSEASL is an arbitration architecture. The notion of sampling past trajectories to compare them against one another in order to provide more stable guidance can be applied to any search algorithm. One condition remains: for the node optimizer block of Figure \ref{fig:tseasl_flowchart} to work as designed, the search space must be a state lattice such that the nodes in the graph can undergo lateral sampling. 
\label{sec:discussion}
\section{Conclusion}
\label{sec:insights}

This paper presents Temporally-Sampled Efficiently Adaptive State Lattices (TSEASL), which is a regional planner architecture that provides the ability for a robot to consider prior trajectories which are updated and/or optimized to reflect changes to the current state and environment. Rather than only considering the optimal trajectory at a single moment in time, which forces a local planner to adjust to potentially drastic changes at the regional planning level, TSEASL presents a module which can select better options for the local planner to follow at the expense of a marginal decrease in regional plan optimality. Results shown in Section \ref{sec:experimental-design} indicate that when compared to the baseline KEASL planner, TSEASL exhibits a greater amount of stability with the trajectories it supplies to the local planner. As a result, this yields more robust navigation performance on the Clearpath Robotics Warthog UGV that it was tested on, evidenced by TSEASL bypassing safety interventions in the same areas where KEASL required them. Quantitatively, experiments were performed to further explore TSEASL's advantages over the baseline, particularly at varying cost threshold percentages ($\alpha$ values). Through the more comprehensive studies that vary $\alpha$, we hope to find a value that minimizes the number of interventions when TSEASL is running live on the robot.

\section{Acknowledgement}
Research was sponsored by the Army Research Laboratory and was accomplished under Cooperative Agreement Numbers W911NF-20-2-0106 and W911NF-24-2-0227. The views and conclusions contained in this document are those of the authors and should not be interpreted as representing the official policies, either expressed or implied, of the DEVCOM Army Research Laboratory or the U.S. Government. The U.S. Government is authorized to reproduce and distribute reprints for Government purposes notwithstanding any copyright notation herein.
\label{sec:thanks}

\bibliography{citation.bib}
\bibliographystyle{plain}
\end{document}